\documentclass{article}
\usepackage{ijcai25}

\usepackage{times}
\usepackage{soul}
\usepackage{url}
\usepackage[hidelinks]{hyperref}
\usepackage[utf8]{inputenc}
\usepackage[small]{caption}
\usepackage{graphicx}
\usepackage{amsmath}
\usepackage{amsthm}
\usepackage{booktabs}
\usepackage{algorithm}
\usepackage{algorithmic}
\usepackage[switch]{lineno}
\usepackage{multirow} 

\title{Adaptive Graph Unlearning}

\author{
Pengfei Ding
\and
Yan Wang\thanks{Corresponding author}\and
Guanfeng Liu\and
Jiajie Zhu\\
\affiliations
Macquarie University, Sydney, Australia\\
\emails
\{pengfei.ding, yan.wang, guanfeng.liu\}@mq.edu.au, jiajie.zhu1@students.mq.edu.au
}

\begin{document}

\maketitle
\begin{abstract}
Graph unlearning, which deletes graph elements such as nodes and edges from trained graph neural networks (GNNs), is crucial for real-world applications where graph data may contain outdated, inaccurate, or privacy-sensitive information. However, existing methods often suffer from (1) incomplete or over unlearning due to neglecting the distinct objectives of different unlearning tasks, and (2) inaccurate identification of neighbors affected by deleted elements across various GNN architectures. To address these limitations, we propose AGU, a novel \textbf{\underline{A}}daptive \textbf{\underline{G}}raph \textbf{\underline{U}}nlearning framework that flexibly adapts to diverse unlearning tasks and GNN architectures. AGU ensures the complete forgetting of deleted elements while preserving the integrity of the remaining graph. It also accurately identifies affected neighbors for each GNN architecture and prioritizes important ones to enhance unlearning performance. Extensive experiments on seven real-world graphs demonstrate that AGU outperforms existing methods in terms of effectiveness, efficiency, and unlearning capability.
\end{abstract}

\section{Introduction}
\begin{figure}[t]
\setlength{\abovecaptionskip}{3pt} 
\setlength{\belowcaptionskip}{-3pt} 
\centering
\scalebox{0.46}{\includegraphics{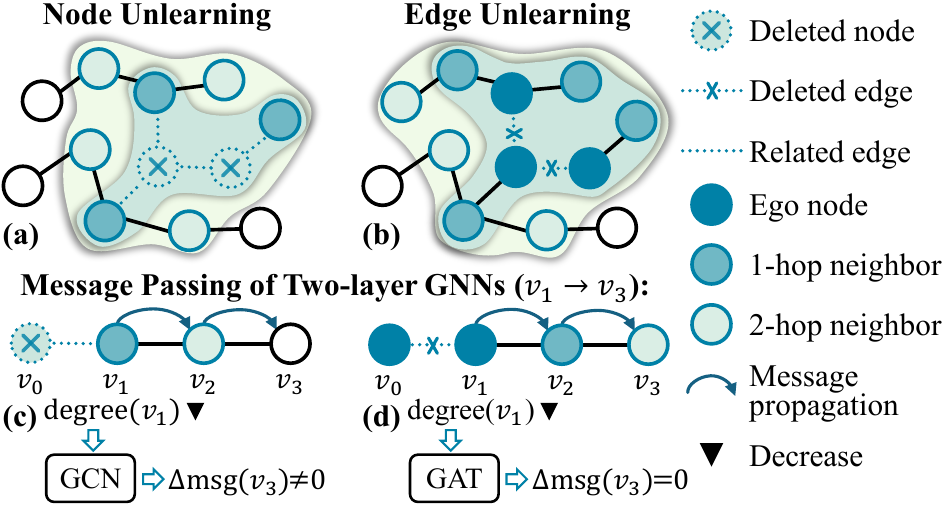}}
\caption{Examples of graph unlearning tasks. The deletion of node $\emph{v}_0$ or edge $(\emph{v}_0, \emph{v}_1)$ affects $\emph{v}_3$ in GCN but not in GAT.}
\label{eg1}
\end{figure}

Machine unlearning (MU) \cite{sekhari2021remember} has emerged as a pivotal step towards responsible AI, enabling the efficient removal of irrelevant, inaccurate, or privacy-sensitive data from trained models. While most MU methods focus on image or text data in Euclidean space, real-world data often appear in the form of graphs, prompting growing interest in unlearning for graph-based models, particularly graph neural networks (GNNs). For instance, fraudulent accounts in financial networks and outdated relations in knowledge graphs should be removed from their corresponding trained GNNs to prevent erroneous predictions. A naive approach is to retrain a new GNN on the updated graph, but this is costly and impractical for frequent unlearning requests. To address this issue, graph unlearning (GU) \cite{fan2025opengu} has been proposed to efficiently eliminate the influence of specific graph elements (e.g., nodes and edges) from trained GNNs. Because the effectiveness of GNNs relies on message propagation across local neighborhoods, effective GU requires not only forgetting information related to the \textit{unlearning elements} (e.g., deleted nodes or edges in Fig. \ref{eg1}), but also minimizing their impact on neighboring nodes, referred to as \textit{affected neighbors} (e.g., ego nodes, 1- and 2-hop neighbors in Fig. \ref{eg1}). 

Recently, a wide range of GU studies \cite{wu2023gif,tan2024unlink} focus on developing general frameworks that can address various GU tasks (e.g., node and edge unlearning) while supporting different backbone GNN architectures (e.g., GCN \cite{kipf2016semi} and GAT \cite{velickovic2017graph}). However, these studies either overlook the distinct objectives of different GU tasks, or neglect the unique message-passing mechanisms of various GNN architectures, leading to two critical limitations as follows.

\textbf{Limitation 1: Ineffective Forgetting of Unlearning Elements.} To remove information about deleted elements from a trained GNN, a series of GU methods \cite{li2024towards,kolipaka2024cognac} adopt a common MU strategy that penalizes the trained GNN for correctly predicting labels associated with deleted elements. However, this strategy is not universally effective across all GU tasks and often results in incomplete or over unlearning: (1) \textit{Incomplete node-unlearning:} These methods input the remaining graph and deleted nodes into the trained GNN to generate labels for unlearning. However, since deleted nodes become isolated after removal, unlearning them in isolation only forgets their intrinsic attributes (e.g., node features), while ignoring the need to forget their original connections (e.g., related edges in Fig. \ref{eg1}(a)). (2) \textit{Over edge-unlearning}: These methods transform edge-level GU tasks into node unlearning scenarios. Specifically, nodes connected to deleted edges (referred to as \textit{ego nodes}, as shown in Fig. \ref{eg1}(b)) are treated as unlearning elements, while the topology of the remaining (undeleted) edges is preserved. However, this transformation inadvertently leads to the unlearning of these ego nodes' features, which should instead be retained in the remaining graph.

\textbf{Limitation 2: Inaccurate Identification of Affected Neighbors.}
Existing GU methods have demonstrated that unlearning performance is heavily influenced by the identification of affected neighbors \cite{wu2023certified,dong2024idea}. However, these methods uniformly set the same affected range for all backbone GNNs. This setting overlooks the fact that different GNN architectures (e.g., GCN and GAT) employ distinct message-passing mechanisms, resulting in different degrees of influence on neighbors in node and edge unlearning. Table \ref{tb1} compares the affected ranges used in existing methods with those identified in our study. 

\begin{table}[t]
\setlength{\abovecaptionskip}{1pt} 
\setlength{\belowcaptionskip}{0pt} 
\centering
\caption{A summary of the affected ranges in recent GU studies. \emph{K} denotes the number of layers in the trained GNN.}
\label{tb1}
\resizebox{85mm}{17.7mm}{
\setlength{\tabcolsep}{3.5mm}{
\begin{tabular}{llll}
\specialrule{0.05em}{0.2pt}{0.2pt}
\textbf{Methods}           & \textbf{Backbone GNNs}      & \textbf{\begin{tabular}[l]{@{}l@{}}Node \\ Unlearn.\end{tabular}} & \textbf{\begin{tabular}[l]{@{}l@{}}Edge \\ Unlearn.\end{tabular}} \\ \specialrule{0.05em}{0.2pt}{0.2pt}
Delete (2023)              & GCN, GAT, GIN      & \emph{K+}1 hops                                                          & \emph{K} hops                                                            \\
CEU (2023)                 & GCN, GIN, SAGE      & -                                                                 & \emph{K} hops                                                            \\
GIF (2023)                 & GCN, SGC, GAT, GIN  & \emph{K+}1 hops                                                          & \emph{K} hops                                                            \\
IDEA (2024)                & GCN, SGC, GAT, GIN & \emph{K+}1 hops                                                          & \emph{K} hops                                                            \\
MEGU (2024)                & GCN, SGC, GAT, GIN & \emph{K} hops                                                            & \emph{K} hops                                                            \\
Cognac (2024)              & GCN, GAT           & \emph{K+}1 hops                                                          & \emph{K} hops                                                            \\ 
ETR (2025)                & GCN, GAT & \emph{K} hops                                                            & \emph{K-}1 hops                                                            \\
\specialrule{0.05em}{0.2pt}{0.2pt}
\multirow{2}{*}{\textbf{AGU (Ours)}} & GCN, SGC            & \emph{K+}1 hops                                                          & \emph{K} hops                                                            \\ \cline{2-4}\rule{0pt}{9pt}
                           & GIN, GAT, SAGE    & \emph{K} hops                                                            & \emph{K-}1 hops                                                          \\ \specialrule{0.05em}{0.2pt}{0.2pt}
\end{tabular}}}

\end{table}

Figs. 1(c) and 1(d) illustrate why the affected ranges differ across various backbone GNNs. Consider the message-passing process of node $\emph{v}_3$ in a two-layer GNN, where $\emph{v}_3$ aggregates messages from its neighbors within two hops (e.g., node features of $\emph{v}_1$ and $\emph{v}_2$). If node $\emph{v}_0$ or edge $(\emph{v}_0, \emph{v}_1)$ is deleted, the degree of $\emph{v}_1$ decreases. Since GCN normalizes neighbor features based on node degrees, this degree change affects the message aggregated by $\emph{v}_3$, i.e., $\Delta\operatorname{msg}(\emph{v}_3)$$\neq$$0$. In contrast, GAT aggregates neighbor information without degree-based normalization, leaving the message aggregated at $\emph{v}_3$ unaffected, i.e., $\Delta\operatorname{msg}(\emph{v}_3)$$=$$0$. Consequently, the affected range in GAT is one hop smaller than that in GCN. Although some GU methods \cite{li2024towards} propose strategies to select important affected neighbors, they overlook the different impact of degree changes on various GNNs, limiting their ability to accurately identify truly affected neighbors.

\textbf{Our Work.} In this paper, we propose an \textbf{\underline{A}}daptive \textbf{\underline{G}}raph \textbf{\underline{U}}nlearning framework (AGU) that flexibly adapts to various GU tasks and backbone GNNs. To address \textbf{Limitation 1}, we propose the \textit{task-adaptive element forgetting}, which contains two basic unlearning modules that can be combined to effectively handle GU tasks at the node, edge, and feature levels. To address \textbf{Limitation 2}, we propose a \textit{GNN-adaptive neighbor selection strategy}, which accurately identifies affected neighbors for each backbone GNN, and prioritizes highly affected neighbors across different GU tasks.

\textbf{Contributions.} (1) \textit{Affected neighbor correction}. 
We identify and rectify inaccuracies in affected neighbor identification, a longstanding issue in the GU literature. (2) \textit{Adaptive GU framework}. We propose a novel framework that flexibly adapts to various GU tasks and backbone GNNs. 
(3) \textit{Universal selection strategy}. We propose a novel neighbor selection strategy that can be integrated into existing GU methods, improving both effectiveness and efficiency.
(4) \textit{SOTA performance}. Extensive experiments on seven real-world graphs show that AGU outperforms state-of-the-art methods in terms of effectiveness, efficiency, and unlearning capability.

\section{Related Work}
\subsection{Graph Neural Networks}
GNNs can be broadly categorized into spectral and spatial approaches. Spectral GNNs define graph convolutions by applying filters to denoise graph signals, with methods like GCN \cite{kipf2016semi} reducing computational complexity through a first-order Chebyshev polynomial approximation. In contrast, spatial GNNs perform convolutions based on node connectivity, updating node representations by aggregating information from neighbors. Since GCN bridges the gap between spectral and spatial GNNs, spatial GNNs, such as GAT \cite{velickovic2017graph} and GIN \cite{xu2018powerful}, have gained popularity due to their efficiency and generality. Further advancements on GNNs can be found in recent surveys \cite{khemani2024review,corso2024graph}.

\subsection{Graph Unlearning} 
Early research on GU primarily focuses on specific unlearning tasks and backbone GNN architectures. For instance, GraphEraser \cite{chen2022graph} and GUIDE \cite{wang2023inductive} partition the graph into multiple shards and process the affected ones, but they only support node unlearning tasks. GraphEditor \cite{cong2022grapheditor}, SGCunlearn \cite{chien2023efficient}, and ScaleGUN \cite{yi2025scalable} offer closed-form solutions with theoretical guarantees but are limited to linear GNNs. Recent advancements \cite{li2024tcgu,yang2024erase} extend GU to a wider range of tasks and backbone GNNs. For example, Delete \cite{cheng2023gnndelete} proposes general operators to remove deleted elements, while GIF \cite{wu2023gif} and IDEA \cite{dong2024idea} estimate parameter changes for various GU tasks. MEGU \cite{li2024towards} improves performance by identifying highly affected neighbors. However, these methods either overlook the distinct objectives of different GU tasks or inaccurately identify affected neighbors, leading to suboptimal performance.

\begin{figure*}[t]
\centering
\scalebox{0.443}{\includegraphics{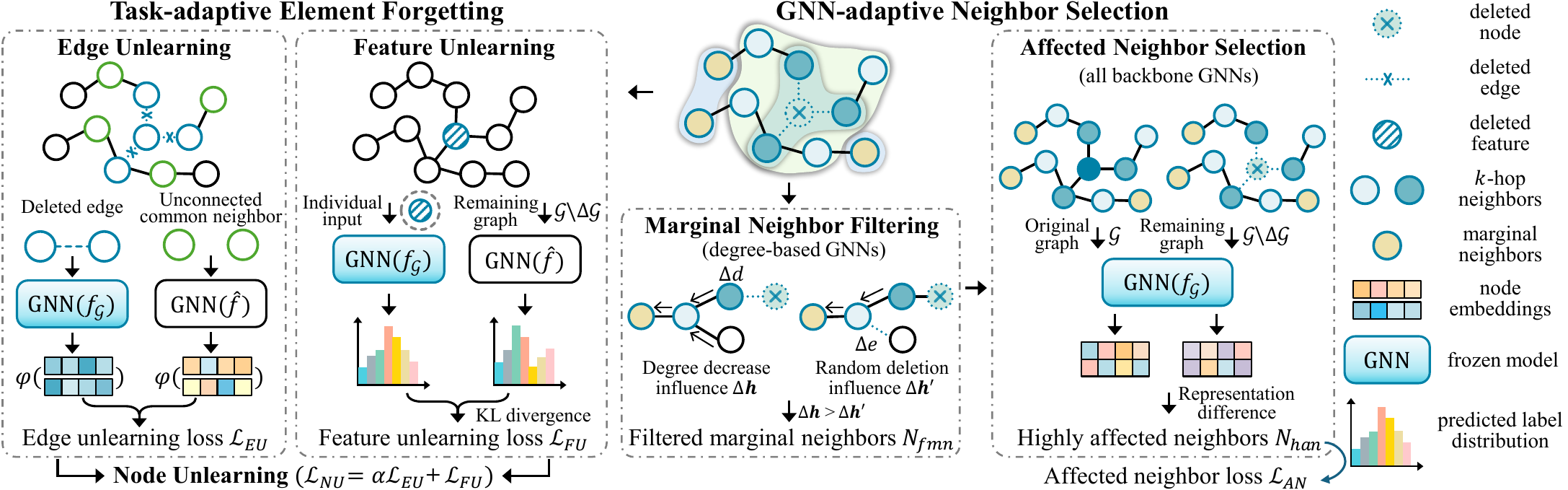}}
\caption{The overall architecture of our proposed AGU framework.}
\label{model}
\end{figure*}

\section{Preliminaries}
\subsection{Notations and Background}
Consider a graph $\mathcal{G}$=$\{\mathcal{V}, \mathcal{E}, \mathcal{X}\}$ with $|\mathcal{V}|$ nodes and $|\mathcal{E}|$ edges. Each node $\emph{v}_\emph{i}$ $\in$ $\mathcal{V}$ is associated with a feature vector $\mathbf{x}_\emph{i}$ $\in$ $\mathcal{X}$. In this paper, we focus on the node classification task. The training set $\mathcal{D}_0$ contains $\emph{N}$ samples $\{\emph{z}_1,$ $ \emph{z}_2, \dots, \emph{z}_\emph{N}\}$, each annotated with a label $\emph{y}$ $\in$ $\mathcal{Y}$. The objective is to train a GNN model $f$ to predict the label of a given node $\emph{v}$ $\in$ $\mathcal{V}$.

\subsection{Graph Unlearning Formalization}
Given a GNN model $f_{\mathcal{G}}$ trained on a graph $\mathcal{G}$, and an unlearning request $\Delta \mathcal{G}$=$\{\Delta \mathcal{V}, \Delta \mathcal{E}, \Delta \mathcal{X}\}$, the goal of GU is to derive a new GNN model $\hat{f}$ that closely approximates $f_{\mathcal{G} \setminus \Delta \mathcal{G}}$, which represents the model retrained from scratch on the remaining graph $\mathcal{G}$$\setminus$$\Delta \mathcal{G}$. GU tasks can be classified into three types: \textbf{Node Unlearning:} $\Delta \mathcal{G}$=$\{\Delta \mathcal{V}, \emptyset, \emptyset\}$, where $\Delta \mathcal{V}$ is the set of nodes to be unlearned. \textbf{Edge Unlearning:} $\Delta \mathcal{G}$=$\{\emptyset, \Delta \mathcal{E}, \emptyset\}$, where $\Delta \mathcal{E}$ is the set of edges to be unlearned. \textbf{Feature Unlearning:} $\Delta \mathcal{G}$=$\{\emptyset, \emptyset, \Delta \mathcal{X}\}$, where $\Delta \mathcal{X}$ denotes the set of features to be unlearned, which will be replaced with zeros. 

\section{Methodology}
\subsection{Framework Overview}
The overall framework of AGU is illustrated in Fig. \ref{model}. To address incomplete or over unlearning across different GU tasks and accurately identify affected neighbors, AGU includes two key components: (1) \textit{Task-adaptive element forgetting}, which contains two basic unlearning modules that can be combined to comprehensively forget information at the node, edge, and feature levels, respectively. (2) \textit{GNN-adaptive neighbor selection}, which determines the accurate affected range for different GNNs and identifies highly affected neighbors. These components empower AGU to perform specific strategies for different GU tasks while effectively mitigating the influence on affected neighbors across various GNN architectures.

\subsection{Task-adaptive Element Forgetting}
GU tasks at the node, edge, and feature levels typically require distinct unlearning strategies, making it complex to address tasks across different levels. However, GU tasks at different levels share commonalities that can help reduce redundant unlearning processes and enable flexible adaptation to complex scenarios. In this section, we design two basic unlearning modules: \textit{edge connection unlearning} and \textit{individual feature unlearning}, which effectively address edge and feature unlearning, respectively. Moreover, their combination provides a comprehensive solution for node unlearning.

\noindent\textbf{Edge Connection Unlearning.} Given a GNN $f_{\mathcal{G}}$ trained on a graph $\mathcal{G}$ and an edge unlearning task $\Delta \mathcal{G}$=$\{\emptyset, \Delta \mathcal{E}, \emptyset\}$, we aim to adjust $f_{\mathcal{G}}$ to obtain an unlearned model $\hat{f}$. Ideally, $\hat{f}$ should be unable to recognize whether a deleted edge $\emph{e}_\emph{uv}$$\in$$\Delta \mathcal{E}$ was originally part of $\mathcal{G}$. A common approach is to use the \textit{deleted edge consistency} loss ($\mathcal{L}_{\operatorname{\emph{DEC}}}$) \cite{cheng2023gnndelete}, which minimizes the difference between the representations of deleted edges and those of randomly selected node pairs:
\begin{equation}
\label{eq1}
    \mathcal{L}_{\operatorname{\emph{DEC}}} = \operatorname{dis} ( \{ \varphi(\hat{\mathbf{h}}_\emph{u},\hat{\mathbf{h}}_\emph{v}) | \emph{e}_\emph{uv}\in\Delta \mathcal{E}\}, \{ \varphi(\mathbf{h}_\emph{p}, \mathbf{h}_\emph{q}) | \emph{p}, \emph{q} \in_\emph{R} \mathcal{V} \} ),
\end{equation}
where $\hat{\mathbf{h}}$ and $\mathbf{h}$ represent the node embeddings output by $\hat{f}$ and $f_{\mathcal{G}}$, respectively. $\varphi(\cdot)$ aggregates the embeddings of two nodes (e.g., via concatenation). $\operatorname{dis}(\cdot)$ is a function positively correlated with the difference. $\in_\emph{R}$ indicates random selection.

However, enforcing the end nodes of edges in $\Delta \mathcal{E}$ to approximate random node pairs may not accurately reflect the actual post-deletion \textit{homophily}, where connected nodes in a graph typically share similar attributes or belong to the same class \cite{zhu2020beyond}. Therefore, even after the removal of edges in $\Delta \mathcal{E}$, their end nodes are expected to retain homophily and should not be treated as randomly paired nodes.

To preserve the underlying homophily between nodes $(\emph{u}, \emph{v})$ $\in$ $\Delta \mathcal{E}$, we refine Eq. (\ref{eq1}) by proposing a new candidate set $\mathcal{V}_{\emph{uv}}^\emph{k}$ for selecting comparable node pairs. Specifically, $\mathcal{V}_{\emph{uv}}^\emph{k}$ consists of common neighbors within the \emph{k}-hop range of $\emph{u}$ and $\emph{v}$, i.e., $\mathcal{V}_{\emph{uv}}^\emph{k}$ = $\mathcal{V}_{\emph{u}}^\emph{k}\cap\mathcal{V}_{\emph{v}}^\emph{k}$, where $\mathcal{V}_{\emph{v}}^\emph{k}$ denotes the set of nodes within \emph{k} hops of $\emph{v}$. We replace $\mathcal{V}$ in Eq. (\ref{eq1}) with $\mathcal{V}_{\emph{uv}}^\emph{k}$ to define our edge unlearning loss, denoted as $\mathcal{L}_{\operatorname{\emph{EU}}}$. If $|\mathcal{V}_{\emph{u}}^\emph{k}\cap\mathcal{V}_{\emph{v}}^\emph{k}|< 2$, we set $\mathcal{V}_{\emph{uv}}^\emph{k}$ as the union of the two sets, i.e., $\mathcal{V}_{\emph{u}}^\emph{k}\cup\mathcal{V}_{\emph{v}}^\emph{k}$. 

\noindent\textbf{Individual Feature Unlearning.} 
Since $f_{\mathcal{G}}$ is trained using node labels as supervision signals, feature unlearning ($\Delta \mathcal{G}$=$\{\emptyset, \emptyset, \Delta \mathcal{X}\}$) requires the unlearned GNN $\hat{f}$ to dissociate the node features in $\Delta \mathcal{X}$ from their corresponding labels predicted by $f_{\mathcal{G}}$. Existing methods \cite{li2024towards,kolipaka2024cognac} typically achieve this through \textit{gradient ascent}, which penalizes $\hat{f}$ for predicting the same labels as $f_{\mathcal{G}}$ for nodes associated with $\Delta \mathcal{X}$.

However, since GNNs aggregate neighbor information to generate representations for prediction, directly applying gradient ascent can lead to the loss of valuable neighbor information, potentially degrading the performance of the unlearned GNN $\hat{f}$. For example, if $f_{\mathcal{G}}$ employs a weighted aggregation strategy that assigns lower weights to the target node $\emph{v}$ but higher weights to its neighbors, $\emph{v}$'s own features contribute less to its prediction. In this case, applying gradient ascent to $\emph{v}$ does not entirely cut the relationship between its features and the corresponding label. Instead, it may inadvertently degrade the prediction performance of other nodes that share similar neighbor information with $\emph{v}$.

To address this issue, we propose a novel feature unlearning strategy that adaptively forgets deleted features while preserving valuable neighbor information. Specifically, we freeze the trained GNN $f_{\mathcal{G}}$ and input individual nodes without edge connections (i.e., $\mathcal{G}$$\setminus$$\mathcal{E}$) into $f_{\mathcal{G}}$ to obtain their independent label distributions $\mathbf{y}^{\prime}$ as self-supervised signals. Then, we enforce the unlearned GNN $\hat{f}$ to output a distinct label distribution $\hat{\mathbf{y}}$ on the remaining graph $\mathcal{G}$$\setminus$$\Delta\mathcal{G}$:
\begin{equation}
\label{eq2}
    \mathcal{L}_{\operatorname{\emph{FU}}} = -\sum\nolimits_{\emph{u} \in \mathcal{V}_{\Delta\mathcal{X}}} \mathcal{L}_{\operatorname{\emph{KL}}} (\mathbf{y}_\emph{u}^\prime, \hat{\mathbf{y}}_\emph{u}),
\end{equation}
where $\mathcal{V}_{\Delta\mathcal{X}}$ denotes the set of nodes whose features are to be unlearned. $\mathcal{L}_{\operatorname{\emph{KL}}}(\cdot)$ is the Kullback-Leibler divergence. The rationale behind Eq. (\ref{eq2}) is: (1) If $f_{\mathcal{G}}$ relies more on $\emph{u}$'s features for prediction, $\mathbf{y}_\emph{u}^\prime$ should approximate $\mathbf{y}_\emph{u}$. Thus, enforcing a distributional difference between $\mathbf{y}_\emph{u}^\prime$ and $\hat{\mathbf{y}}_\emph{u}$ encourages $\hat{f}$ to forget the correlation between $\emph{u}$'s features and the prediction made by $f_{\mathcal{G}}$. (2) If $f_{\mathcal{G}}$ relies more on $\emph{u}$'s neighbor information for prediction, $\mathbf{y}_\emph{u}^\prime$ should differ from $\mathbf{y}_\emph{u}$. Thus, enforcing a distributional difference between $\mathbf{y}_\emph{u}^\prime$ and $\hat{\mathbf{y}}_\emph{u}$ can forget $\emph{u}$'s own features while preserving valuable neighbor information. 

\noindent\textbf{Node Unlearning.} 
A simple yet common strategy to forget deleted nodes $\Delta\mathcal{V}$ in a trained node classification model $f_{\mathcal{G}}$ is to apply a reverse cross-entropy loss: $-\sum_{\emph{u} \in \Delta \mathcal{V}} \mathcal{L}_{\operatorname{\emph{CE}}} ( \emph{y}_\emph{u}, \hat{\emph{y}}_\emph{u})$, where $\emph{y}_\emph{u}$ is the prediction of $f_{\mathcal{G}}$ on $\mathcal{G}$, and $\hat{\emph{y}}_\emph{u}$ is the prediction of $\hat{f}$ on the remaining graph $\mathcal{G}$$\setminus$$\Delta \mathcal{V}$. However, since the nodes in $\Delta \mathcal{V}$ become isolated after deletion, $\hat{\emph{y}}_\emph{u}$ depends only on $\emph{u}$'s own features and is no longer influenced by its original neighbors. Thus, applying this loss function cannot ensure $\hat{f}$ fully forgets the original connections between $\emph{u}$ and its neighbors.

For comprehensive node unlearning, we adopt a new perspective by transforming the task from $\{\Delta \mathcal{V}, \emptyset, \emptyset\}$ to $\{\emptyset, \Delta \mathcal{E}, $ $\Delta \mathcal{X}\}$, where $\Delta \mathcal{E}$ represents the edges connected to nodes in $\Delta \mathcal{V}$, and $\Delta \mathcal{X}$ denotes node features of $\Delta \mathcal{V}$. This transformation allows us to combine our proposed edge and feature unlearning modules to derive the node unlearning loss $\mathcal{L}_{\operatorname{\emph{NU}}}$:
\begin{equation}
\label{eq3}
    \mathcal{L}_{\operatorname{\emph{NU}}} = \alpha\mathcal{L}_{\operatorname{\emph{EU}}} + \mathcal{L}_{\operatorname{\emph{FU}}},
\end{equation}
where $\alpha$ is a loss coefficient hyper-parameter.

\subsection{GNN-adaptive Neighbor Selection}
Existing studies \cite{wu2023certified,li2024towards} show that unlearning performance heavily depends on the identification of affected neighbors. Moreover, selecting highly affected neighbors can improve both the effectiveness and efficiency of unlearning. In this section, we address inaccuracies in existing methods for identifying affected neighbors and propose a novel GNN-adaptive selection strategy.

\noindent\textbf{Accurate Affected Neighbor Identification.}  
Given a $\emph{k}$-layer trained GNN $f_{\mathcal{G}}$ on graph $\mathcal{G}$ and an unlearning request $\Delta\mathcal{G}$, existing GU methods identify the affected neighbors $\mathcal{N}_\emph{aff}$ either directly based on the number of layers (i.e., $\emph{k}$) \cite{dong2024idea,li2024towards} or through $\emph{k}$-step normalized propagation \cite{wu2023gif,chen2023characterizing} as follows:
\begin{gather}
\mathcal{N}_\emph{aff} = \left\{ \emph{i} \mid \Delta \mathbf{H}(\hat{\mathbf{A}}, \mathbf{A})_\emph{i} \neq 0, \ \emph{i} \in \mathcal{G}\setminus\Delta\mathcal{G} \right\},
\\
\Delta \mathbf{H}(\hat{\mathbf{A}}, \mathbf{A}) = [ (\hat{\mathbf{D}}^{-\frac{1}{2}} \hat{\mathbf{A}} \hat{\mathbf{D}}^{-\frac{1}{2}})^\emph{k} - ({\mathbf{D}}^{-\frac{1}{2}} {\mathbf{A}} {\mathbf{D}}^{-\frac{1}{2}})^\emph{k}] \mathbf{X},
\end{gather}
where $\mathbf{A}$ and $\hat{\mathbf{A}}$ are the adjacency matrices of $\mathcal{G}$ and $\mathcal{G}$$\setminus$$\Delta\mathcal{G}$, respectively, with $\mathbf{D}$ and $\hat{\mathbf{D}}$ as their corresponding degree matrices. $\Delta \mathbf{H}(\hat{\mathbf{A}}, \mathbf{A})$ quantifies the propagation change before and after deleting unlearning elements. Nodes outside $\Delta\mathcal{G}$ that satisfy $\Delta \mathbf{H}(\hat{\mathbf{A}}, \mathbf{A})$$\neq$$0$ are considered affected neighbors.

However, this identification method is limited to backbone GNNs that incorporate node degrees in message passing (e.g., GCN, SGC \cite{wu2019simplifying}), which we refer to as \textit{degree-based} GNNs. In contrast, \textit{degree-free} GNNs (e.g., GAT, GIN \cite{xu2018powerful}) aggregate neighbor information without relying on node degrees, leading to a different set of affected neighbors. As discussed in Section 1, degree-free GNNs typically influence fewer nodes than their degree-based counterparts. We validate this difference through experiments (see Appendix \cite{ding2025agu} for details). This influence discrepancy suggests that existing methods incorrectly set the same range of affected neighbors for all backbone GNNs. To address this issue, we summarize the correct affected range for popular GNNs in Table \ref{tb1}, and propose a simple yet general strategy to accurately identify affected neighbors without analyzing the architecture of the trained GNN $f_\mathcal{G}$:
\begin{equation}
\mathcal{N}_\emph{ac} = \left\{\emph{v} \mid f_\emph{rand}(\mathcal{G})_\emph{v} \neq f_\emph{rand}(\mathcal{G}\setminus\Delta\mathcal{G})_\emph{v}, \emph{v} \in \mathcal{G}\setminus\Delta\mathcal{G}\right\},  
\end{equation}
where $f_\emph{rand}$ denotes a randomly initialized GNN with same architecture as $f_\mathcal{G}$. Note that in practice, it is crucial to remove randomness from $f_\emph{rand}$'s message-passing process to ensure accurate identification of affected neighbors. 

\begin{table*}[t]
\setlength{\abovecaptionskip}{2pt}
\setlength{\belowcaptionskip}{-1pt}
\centering
\caption{Performance comparison in terms of F1 score. The best results are highlighted in \textbf{bold}, while the second-best results are \underline{underlined}.}
\label{exp1}
\resizebox{176mm}{34mm}{
\setlength{\tabcolsep}{1.8mm}{
\begin{tabular}{cc|cc|cc|cc|cc|cc|cc|cc}
\specialrule{0.05em}{1pt}{1pt} 
\multirow{2}{*}{{Task}}  & {Method}  & \multicolumn{2}{c|}{{Cora}}            & \multicolumn{2}{c|}{{Citeseer}}        & \multicolumn{2}{c|}{{PubMed}}          & \multicolumn{2}{c|}{{Photo}}           & \multicolumn{2}{c|}{{Computer}}        & \multicolumn{2}{c|}{{CS}}              & \multicolumn{2}{c}{{Flickr}}          \\
                       & {Bone}    & {GCN}              & {GIN}              & {GCN}              & {GIN}              & {GCN}              & {GIN}              & {GCN}              & {GIN}              & {GCN}              & {GIN}              & {GCN}              & {GIN}              & {GCN}              & {GIN}              \\\specialrule{0.05em}{1pt}{1pt}
\multirow{9}{*}{{Node}}  & Retrain & 86.1±.4          & 83.5±.7          & 74.6±.3          & 73.6±.7          & 88.6±.7          & 84.9±.4          & 91.8±2           & 85.9±.4          & 83.7±.7          & 77.8±.8          & 92.1±.2          & 90.9±.3          & 49.1±.1          & 47.3±.4          \\
                       & Delete  & 82.5±.5          & 76.2±1           & 71.4±.5          & 67.4±1           & 84.6±.2          & 76.1±3           & 89.2±.4          & 83.5±.0          & 84.2±.5          & 75.2±.0          & 91.4±.3          & 87.9±1           & 45.5±.4          & 40.4±.4          \\
                       & GIF     & 84.5±.6          & 82.1±.7          & 73.2±.5          & 72.4±.7          & 86.5±.9          & 84.2±.3          & 89.7±.8          & 76.2±.6          & 83.5±.4          & 77.1±.3          & 91.7±.3          & 88.7±.4          & 47.4±.8          & 42.5±.5          \\
                       & IDEA    & 80.2±2           & 82.3±.7          & 69.7±2           & 72.6±.7          & 82.3±3           & 84.3±.3          & 88.1±2           & 78.8±.4          & 83.3±.4          & 78.2±.2          & 91.6±.3          & 88.4±.4          & 47.1±.2          & 42.5±.6          \\
                       & MEGU    & \underline{85.4±.6}          & \underline{83.6±.3}          & \underline{74.8±.5}          & 73.1±.7          & 87.1±.2          & 84.9±.3          & 91.9±.4          & 80.6±.4          & \underline{85.8±.3}          & 78.3±.2          & \underline{92.2±.3}          & 89.4±.4          & \underline{49.3±.2}          & \underline{44.6±.3}          \\
                       & UTU     & 84.2±.0          & 80.5±.0          & 72.7±.0          & 70.6±.0          & 86.7±.0          & 83.6±.0          & 89.5±.0          & 83.8±.0          & 80.8±.0          & 77.3±.0          & 91.4±.0          & 88.3±.0          & 47.1±.0          & 41.8±.0          \\
                       & ETR     & 84.9±.7          & 83.4±.8          & 74.2±.6          & \underline{73.3±.6}          & 87.3±.9          & 85.1±.3          & \underline{92.3±.3}          & 74.2±.6          & 85.7±.3          & 70.6±.2          & 91.8±.2          & 90.4±.5          & 48.1±.8          & 40.2±.2          \\
                       & Cognac  & 85.1±.5          & 82.9±.8          & 74.0±.6          & 73.2±.9          & \underline{87.4±.1}          & \underline{85.2±.1}          & 92.2±.2          & \underline{84.1±.4}          & 85.2±.5          & \underline{78.4±.3}          & 92.1±.1          & \underline{90.9±.2}          & 48.7±.1          & 43.4±.3          \\
                       & AGU     & \textbf{85.9±.3} & \textbf{84.3±.2} & \textbf{76.2±.5} & \textbf{73.8±.7} & \textbf{88.4±.2} & \textbf{85.5±.3} & \textbf{92.6±.1} & \textbf{85.7±.3} & \textbf{86.3±.4} & \textbf{79.3±.5} & \textbf{92.8±.3} & \textbf{91.7±.4} & \textbf{50.2±.2} & \textbf{46.8±.5} \\\specialrule{0.05em}{1pt}{1pt}
\multirow{11.3}{*}{{Edge}} & {Bone}    & {SAGE}             & {GAT}              & {SAGE}             & {GAT}              & {SAGE}             & {GAT}              & {SAGE}             & {GAT}              & {SAGE}             & {GAT}              & {SAGE}             & {GAT}              & {SAGE}             & {GAT}              \\\specialrule{0.05em}{1pt}{1pt}
                       & Retrain & 86.8±.3          & 86.5±.2          & 78.0±.6          & 75.1±.2          & 89.1±.4          & 84.9±.4          & 94.3±2           & 90.1±.2          & 88.4±.8                & 84.6±.7          & 94.1±.2          & 91.3±.3          & 49.8±.8                & 49.2±.4          \\
                       & Delete  & 84.7±.4          & 84.3±.7          & 73.7±.5          & 73.2±.1          & 88.9±.3          & 82.6±.9          & 92.6±.6          & 85.4±.4          & 86.5±.6                & 78.7±.4          & 92.8±.4          & 91.1±.2          & 47.2±.7                & 46.2±.5          \\
                       & GIF     & \underline{86.1±.1}          & 85.2±.4          & 77.6±.5          & 73.7±.7          & \underline{89.3±.0}          & 85.4±.3          & 93.6±.6          & 87.8±.7          & 86.3±.4                & 78.2±.3          & 93.6±.1          & 90.7±.2          & 48.7±.4                & 46.7±.3          \\
                       & IDEA    & 85.6±.7          & 85.2±.4          & 76.7±.5          & 73.8±.7          & 88.7±.3          & 85.2±.3          & 91.7±.5          & 87.6±.7          & 84.5±.7              & 78.3±.7          & 92.5±.2          & 90.9±.2          & 47.8±.8                & 46.7±.5          \\
                       & MEGU    & 85.9±.3          & \underline{86.2±.1}          & \underline{78.1±.1}          & 74.1±.8          & 89.2±.3          & 86.4±.6          & \underline{94.3±.2}          & 89.7±.3          & 87.5±.1                & 82.8±.4          & \underline{94.2±.1}          & \underline{92.1±.3}          & \underline{49.6±.3}                & 45.8±.6          \\
                       & UTU     & 83.8±.0          & 82.2±.0          & 77.5±.0          & 71.6±.0          & 88.2±.0          & 80.3±.0          & 93.7±.0          & 85.9±.0          & 87.1±.0                & 73.7±.0          & 94.1±.0          & 89.9±.0          & 47.5±.0                & 45.1±.0          \\
                       & ETR     & 84.2±.2          & 85.5±.5          & 73.2±.3          & \underline{74.3±.3}          & 87.5±.2          & \underline{86.4±.1}          & 94.1±.6          & \underline{90.9±.4}          & 87.3±.4               & \underline{83.5±.3}          & {94.1±.2}          & 91.1±1           & 49.5±,6                & 47.1±.7          \\
                       & Cognac  & 83.9±.6          & 85.9±.6          & 77.9±.5          & 73.6±.4          & 88.4±.2          & 86.3±.2          & 92.5±.2          & 90.4±.5          & \underline{87.6±.3}                & 81.9±.5          & 93.9±.2          & 91.8±.2          & 49.4±.2                & \underline{47.6±.2}          \\
                       & AGU     & \textbf{87.4±.2} & \textbf{86.9±.1} & \textbf{78.6±.2} & \textbf{75.5±.3} & \textbf{90.2±.3} & \textbf{87.1±.4} & \textbf{95.3±.5} & \textbf{92.3±.6} & \textbf{88.7±.4}       & \textbf{85.4±.3} & \textbf{94.5±.2} & \textbf{92.7±.2} & \textbf{50.3±.5}       & \textbf{48.6±.1}\\\specialrule{0.05em}{1pt}{1pt}
\end{tabular}}}
\end{table*}

\noindent\textbf{Marginal Neighbor Filtering.}
For node and edge unlearning, degree-based GNNs affect an additional hop of neighbors compared to degree-free GNNs (as shown in Table \ref{tb1}). We refer to these extra affected neighbors as \textit{marginal neighbors}. Note that for feature unlearning, since the graph structure remains unchanged after unlearning, the affected range is the same for all backbone GNNs.

We first analyze the impact of unlearning on marginal neighbors compared to other affected neighbors. Consider the message-passing process in a simple and classic degree-based GNN, SGC. The output for node $\emph{i}$ at the $\emph{l}$-th layer is:
\begin{equation}
\label{sgc}
\mathbf{h}_\emph{i}^{\emph{(l)}} = \sum\nolimits_{\emph{j} \in \mathcal{N}_\emph{i}}  \frac{\mathbf{h}_\emph{j}^{\emph{(l-}1\emph{)}} }{\sqrt{\emph{d}_\emph{i}\emph{d}_\emph{j}}}= \sum\nolimits_{\emph{j} \in \mathcal{N}^l_\emph{i}} \frac{\mathbf{x}_\emph{j}}{\prod_{t\emph{=}1}^\emph{l} \sqrt{\emph{d}_{\emph{j}_{t\emph{-}1}} \emph{d}_{\emph{j}_t}}},
\end{equation} 
where $\mathcal{N}^l_\emph{i}$ denotes the set of $\emph{i}$'s neighbors at the $\emph{l}$-th hop. $\emph{d}_\emph{i}$ is the degree of node $\emph{i}$. The term $\prod_{t\emph{=}1}^\emph{l} \sqrt{\emph{d}_{\emph{j}_{t\emph{-}1}} \emph{d}_{\emph{j}_t}}$ represents the degree normalization over $\emph{l}$ layers.

In node unlearning, if $\emph{i}$ is a marginal neighbor, the deletion of unlearning nodes causes only a minor degree change, $\Delta \emph{d}$, for some of $\emph{i}$'s neighbors $\emph{l}$ hops away (i.e., $\emph{d}_{\emph{j}_l}$), while leaving $\mathcal{N}^l_\emph{i}$ unchanged. In contrast, for other affected neighbors, message passing is often obstructed due to a reduction in $\mathcal{N}^l_\emph{i}$, leading to a more significant change in $\Delta \mathbf{h}$. Therefore, marginal neighbors tend to be less influenced than other affected nodes. To filter out marginal neighbors with minimal impact from degree changes, we propose the following filtering strategy: \textit{A marginal neighbor should be considered only if the degree change in its neighbors exceeds the effect of randomly removing one of its neighbors:}
\begin{equation}
    \mathcal{N}_\emph{fmn} = \{\emph{i}|\lvert\Delta \mathbf{H}(\hat{\mathbf{A}}, \mathbf{A})_\emph{i}\rvert-\lvert\Delta \mathbf{H}({\mathbf{A}^\prime}, \mathbf{A})_\emph{i}\rvert >\theta\},
\end{equation}
where $\mathbf{A}^\prime$ is the adjacency matrix obtained after randomly deleting an edge within the \emph{k}-hop neighborhood of each node in $\Delta \mathcal{G}$. $\theta$ represents the difference threshold parameter.

\noindent\textbf{Affected Neighbor Selection.} In this part, we propose a general strategy to select important affected neighbors. Existing GU methods typically employ \textit{inverse feature propagation} on the original graph $\mathcal{G}$ to identify highly affected neighbors. Specifically, they perform feature propagation (e.g., ${\mathbf{A}}^\emph{k}\mathbf{X}$) using the original feature matrix $\mathbf{X}$ and its inverse counterpart $\mathbf{X}^\prime$=$\mathbf{1}$-$\mathbf{X}$. By comparing propagation results for the same node, they identify highly affected neighbors as those with significant differences. However, this approach overlooks variations across different GU tasks and GNN architectures.

To address this limitation, we propose a task- and GNN- adaptive neighbor selection module. The idea is to utilize the trained GNN to directly identify important affected neighbors. Specifically, we freeze the trained model $f_{\mathcal{G}}$ and input $\mathcal{G}$ and $\mathcal{G}$$\setminus$$\Delta\mathcal{G}$ into $f_{\mathcal{G}}$. We then compare the differences in the node representations output by $f_{\mathcal{G}}$ for these two graphs:
\begin{equation}
\label{han}
    \operatorname{diff}(\emph{v}) = \operatorname{dis}(f_{\mathcal{G}}(\mathcal{G}\setminus\Delta\mathcal{G})_\emph{v},f_{\mathcal{G}}(\mathcal{G})_\emph{v}),
\end{equation}
where $f_{\mathcal{G}}(\cdot)_\emph{v}$ denotes the representation of node $\emph{v}$ output by $f_{\mathcal{G}}$. $\operatorname{dis}(\cdot)$ is a function that positively correlates with the difference. Intuitively, nodes with higher $\operatorname{diff}(\cdot)$ values are more significantly affected by the unlearning elements. Thus, we select the top-$\emph{k}_\emph{ans}$ nodes with the highest $\operatorname{diff}(\cdot)$ values as important affected neighbors, denoted as $\mathcal{N}_\emph{han}$. For degree-based GNNs, we first apply marginal neighbor filtering, then use the remaining affected neighbors for further selection.

\noindent\textbf{Optimization Objective.} 
Unlike the labels of unlearning elements in $\Delta\mathcal{G}$, which are intended to be forgotten in the unlearned GNN $\hat{f}$, the predictive accuracy for the highly affected neighbors $\mathcal{N}_\emph{han}$ should be preserved in $\hat{f}$. To this end, we follow \cite{li2024towards} and treat the predictions of the original model ($f_{\mathcal{G}}$) as self-supervised signals for $\mathcal{N}_\emph{han}$. Specifically, we freeze the parameters of $f_{\mathcal{G}}$ and define a cross-entropy loss $\mathcal{L}_{\operatorname{\emph{AN}}}$$=$$\sum_{\emph{u} \in \mathcal{N}_\emph{han}} \mathcal{L}_{\operatorname{\emph{CE}}} ( \emph{y}_\emph{u}, \hat{\emph{y}}_\emph{u})$, where $\emph{y}_\emph{u}$ is the prediction of $f_{\mathcal{G}}$ on $\mathcal{G}$, and $\hat{\emph{y}}_\emph{u}$ is the prediction of $\hat{f}$ on $\mathcal{G}$$\setminus$$\Delta \mathcal{G}$. 

Finally, by combining $\mathcal{L}_{\operatorname{\emph{AN}}}$ with the loss on the unlearning elements $\Delta \mathcal{G}$, the overall loss is formulated as:
\begin{equation}
        \mathcal{L} = \mathcal{L}_{\operatorname{\emph{EF}}} + \mathcal{L}_{\operatorname{\emph{AN}}},
\end{equation}
where $\mathcal{L}_{\operatorname{\emph{EF}}}$ is selected from $\{\mathcal{L}_{\operatorname{\emph{NU}}}, \mathcal{L}_{\operatorname{\emph{EU}}}, \mathcal{L}_{\operatorname{\emph{FU}}}\}$ based on the specific GU task.

\section{Experiments}
We conduct extensive experiments to answer the following research questions: \textbf{RQ1:} How does AGU perform compared to state-of-the-art methods? \textbf{RQ2:} How does each proposed module in AGU contribute to the overall performance? \textbf{RQ3:} Can our proposed neighbor selection strategies enhance the performance of existing methods? \textbf{RQ4:} How do different parameter settings influence AGU's performance? 

\subsection{Experimental Setup}
\textbf{Datasets.} We select seven widely used datasets: Cora, Citeseer, PubMed \cite{yang2016revisiting}, Amazon-Photo, Amazon-Computers, Coauthor-CS \cite{shchur2018pitfalls}, and Flickr \cite{zenggraphsaint}. The datasets are split following the guidelines of recent GU studies \cite{cheng2023gnndelete,li2024towards}, with $80\%$ of the nodes used for training and $20\%$ for testing. Detailed statistics and descriptions of these datasets are provided in the Appendix \cite{ding2025agu}.

\noindent\textbf{Backbone GNNs and Baselines.} We evaluate the adaptability of AGU using two \textbf{degree-based GNNs:} GCN \cite{kipf2016semi} and SGC \cite{wu2019simplifying}, and three \textbf{degree-free GNNs:} GAT \cite{velickovic2017graph}, SAGE \cite{hamilton2017inductive}, and GIN \cite{xu2018powerful}. We compare AGU against \textbf{seven state-of-the-art baselines:} Delete \cite{cheng2023gnndelete}, GIF \cite{wu2023gif}, IDEA \cite{dong2024idea}, MEGU \cite{li2024towards}, UTU \cite{tan2024unlink}, ETR \cite{yang2024erase}, and Cognac \cite{kolipaka2024cognac}. Detailed descriptions of these GNNs and baselines are provided in the Appendix. For all methods, we set the embedding dimension to $64$, and fix the number of GNN layers at $2$. Baseline parameters are initialized using the values reported in the original papers and further fine-tuned for optimal performance. In AGU, the edge unlearning loss $\mathcal{L}_\emph{EU}$ uses concatenation for $\varphi(\cdot)$ and mean-squared error for $\operatorname{dis}(\cdot)$, while Eq. (\ref{han}) uses cosine similarity for $\operatorname{dis}(\cdot)$. The loss coefficient parameter $\alpha$ is set to $0.1$. To ensure a fair comparison, each experiment is repeated $10$ times and we report the average performance. 

\noindent\textbf{Unlearning Tasks.} In our experiments, GU tasks are designed as follows: (1) \textbf{Node-level:} We randomly delete $5\%$ of nodes from the training graph and remove their associated edges. (2) \textbf{Edge-level:} We randomly delete $5\%$ of edges from the training graph. (3) \textbf{Feature-level:} We randomly select $5\%$ of nodes from the training graph and mask their full-dimensional features. Results for other unlearn ratios are provided in the Appendix. All methods are evaluated using the Micro-F1 score for node classification. To further evaluate the unlearning capability of GU methods, we adopt the widely used \textit{Edge Attack} approach \cite{wu2023gif,li2024towards}. Specifically, we add noisy edges into the training graph to mislead representation learning. Each added edge connects two nodes from different classes to maximize its adversarial effect. These noisy edges are then treated as unlearning elements. The unlearning capability of GU methods is evaluated by their ability to mitigate the negative impact of these noisy edges on predictive performance. 

\begin{table}[t]
\setlength{\abovecaptionskip}{1pt}
\setlength{\belowcaptionskip}{0pt}
\centering
\caption{Performance comparison of feature unlearning.}
\label{exp2}
\resizebox{84mm}{31mm}{
\setlength{\tabcolsep}{1.8mm}{
\begin{tabular}{cc|cccccc}
\specialrule{0.05em}{1.5pt}{1.5pt}
\multirow{2}{*}{Bone} & \multirow{2}{*}{Method} & \multicolumn{2}{c}{Cora}         & \multicolumn{2}{c}{Citeseer}     & \multicolumn{2}{c}{PubMed}       \\
                      &                         & F1               & Time             & F1               & Time             & F1               & Time             \\\specialrule{0.05em}{1.5pt}{1.5pt}
\multirow{8}{*}{SGC}  & Retrain                 & 82.4±.1          & 5.49          & 72.2±.1          & 7.38          & 83.5±.1          & 24.1          \\
                      & Delete                  & 81.6±.2          & 1.75          & 70.4±.3          & 1.18          & 82.5±.1          & 2.96          \\
                      & GIF                     & 82.6±.2          & 0.23          & 70.2±.2          & 0.24          & 83.1±.1          & 1.36          \\
                      & IDEA                    & 83.8±.4           & 0.24          & 61.5±.7           & 0.24          & 82.3±.9          & 1.42          \\
                      & MEGU                    & \underline{84.3±.2}          & 0.14          & \underline{73.1±.2}          & 0.18          & 82.9±.1          & 0.27          \\
                      & ETR                     & 83.9±.4          & \textbf{0.01} & 72.8±.1          & \textbf{0.02} & \underline{84.1±.4}          & \textbf{0.03} \\
                      & Cognac                  & 83.4±.2          & 0.84          & 72.9±.2          & 0.91          & 83.9±.1          & 1.63          \\
                      & AGU                     & \textbf{85.1±.2} & \underline{0.07}          & \textbf{73.8±.2} & \underline{0.12}          & \textbf{84.7±.1} & \underline{0.13}          \\\specialrule{0.05em}{1.5pt}{1.5pt}
\multirow{8}{*}{GIN}  & Retrain                 & 84.5±.2          & 10.5          & 72.3±.4          & 16.3          & 84.1±.4          & 42.6          \\
                      & Delete                  & 83.4±.3          & 2.53          & 68.4±1           & 2.36          & 76.8±3           & 2.72          \\
                      & GIF                     & 85.4±.1          & 0.26          & 72.2±.1          & 0.26          & 84.2±.3          & 2.30          \\
                      & IDEA                    & 85.3±.4          & 0.25          & 71.9±.7          & 0.26          & 84.2±.7          & 2.22          \\
                      & MEGU                    & \underline{85.6±.2}          & 0.15          & 72.9±.3          & 0.15          & 84.9±.1          & 0.24          \\
                      & ETR                     & 84.3±.3          & \textbf{0.01} & \underline{73.2±.2}          & \textbf{0.02} & 84.7±.3          & \textbf{0.03} \\
                      & Cognac                  & 84.3±.4          & 0.83          & 71.6±.8          & 0.69          & \underline{85.1±.2}          & 1.76          \\
                      & AGU                     & \textbf{86.7±.2} & \underline{0.07}          & \textbf{73.9±.4} & \underline{0.08}          & \textbf{85.5±.3} & \underline{0.08}  \\\specialrule{0.05em}{1.5pt}{1.5pt}       
\end{tabular}}}
\end{table}

\subsection{Performance Comparison (RQ1)}
We compare the performance of all methods from three perspectives: \textit{effectiveness}, \textit{efficiency}, and \textit{unlearning capability}.

\noindent\textbf{Effectiveness.}
Tables \ref{exp1} and \ref{exp2} compare the node classification performance across various GU tasks and backbone GNNs. Notably, AGU consistently outperforms all baselines in F1-score across all settings. On average, AGU achieves a $1.37\%$ improvement over the best-performing baseline in node unlearning (ranging from $0.35\%$ to $4.93\%$), $1.21\%$ in edge unlearning ($0.31\%$ to $2.28\%$), and $1.26\%$ in feature unlearning ($0.47\%$ to $3.31\%$). These results demonstrate AGU's robustness across diverse GU tasks and backbone GNNs, highlighting the effectiveness of its task-adaptive element forgetting and GNN-adaptive neighbor selection. In contrast, existing methods adopt uniform unlearning strategies, leading to inferior overall performance. For instance, Delete, which adopts an edge-based unlearning strategy, performs well in edge-level GU tasks but struggles in other GU tasks. These findings further validate the necessity of designing task-specific unlearning strategies for different GU tasks.

\begin{table}[t]
\setlength{\abovecaptionskip}{-7pt}
\setlength{\belowcaptionskip}{0pt}
\centering
\caption{Comparison of node unlearning runtime.}
\label{exp3}
\resizebox{84mm}{30mm}{
\setlength{\tabcolsep}{1.2mm}{
\begin{tabular}{cc|ccccccc}
\\\specialrule{0.05em}{1.5pt}{1.5pt}       
Bone                 & Method   & Cora          & Citeseer      & PubMed        & Photo         & Computer      & CS            & Flickr        \\\specialrule{0.05em}{1.5pt}{1.5pt}       
\multirow{9}{*}{GCN} & Retrain  & 8.47          & 13.5          & 35.7          & 21.6          & 49.3          & 74.2          & 172           \\ \cline{2-9} \rule{0pt}{9pt}
                     & GIF      & 0.29          & 0.26          & 0.56          & 0.49          & 0.29          & 0.31          & 1.22          \\
                     & IDEA     & 0.27          & 0.26          & 0.55          & 0.49          & 0.27          & 0.28          & 1.19          \\
                     & ETR      & \textbf{0.02} & \textbf{0.03} & \textbf{0.05} & \textbf{0.03} & \textbf{0.06} & \underline{0.14}          & \textbf{0.22} \\ \cline{2-9} \rule{0pt}{9pt}
                     & Delete   & 2.14          & 2.01          & 2.57          & 2.18          & 2.26          & 2.74          & 8.32          \\
                     & Cognac   & 1.01          & 0.83          & 2.15          & 1.27          & 3.34          & 4.71          & 4.33          \\
                     & MEGU     & 0.18          & 0.16          & 0.25          & 0.22          & 0.25          & 0.26          & 1.82          \\
                     & AGU      & \underline{0.11}          & \underline{0.09}          & \underline{0.10}          & \underline{0.11}          & \underline{0.10}          & \textbf{0.12}          & \underline{0.31}          \\\specialrule{0.05em}{1pt}{1pt}    
\multirow{9}{*}{GAT} & Retrain  & 13.8          & 17.1          & 47.2          & 23.1          & 50.1          & 77.4          & 196           \\ \cline{2-9} \rule{0pt}{9pt}
                     & GIF      & 0.69          & 0.68          & 1.47          & 0.93          & 2.73          & 0.71          & 1.14          \\
                     & IDEA     & 0.65          & 0.65          & 1.46          & 0.95          & 2.69          & 0.63          & 1.11          \\
                     & ETR      & \textbf{0.04} & \textbf{0.04} & \textbf{0.06} & \textbf{0.03} & \textbf{0.08} & \textbf{0.12} & \underline{0.26}          \\ \cline{2-9} \rule{0pt}{9pt}
                     & Delete   & 3.23          & 2.62          & 3.59          & 2.32          & 2.87          & 2.77          & 9.13          \\
                     & Cognac   & 1.57          & 1.38          & 2.42          & 1.54          & 3.83          & 6.85          & 2.21          \\
                     & MEGU     & 0.30          & 0.30          & 0.44          & 0.32          & 0.36          & 0.37          & 0.98          \\
                     & AGU      & \underline{0.16}          & \underline{0.15}          & \underline{0.18}          & \underline{0.13}          & \underline{0.15}          & \underline{0.17}          & \textbf{0.24}          \\\specialrule{0.05em}{1.5pt}{1.5pt}
\end{tabular}}}
\end{table}

\begin{figure}[t]
\setlength{\abovecaptionskip}{1pt}
\setlength{\belowcaptionskip}{0pt}
\centering
\scalebox{0.292}{\includegraphics{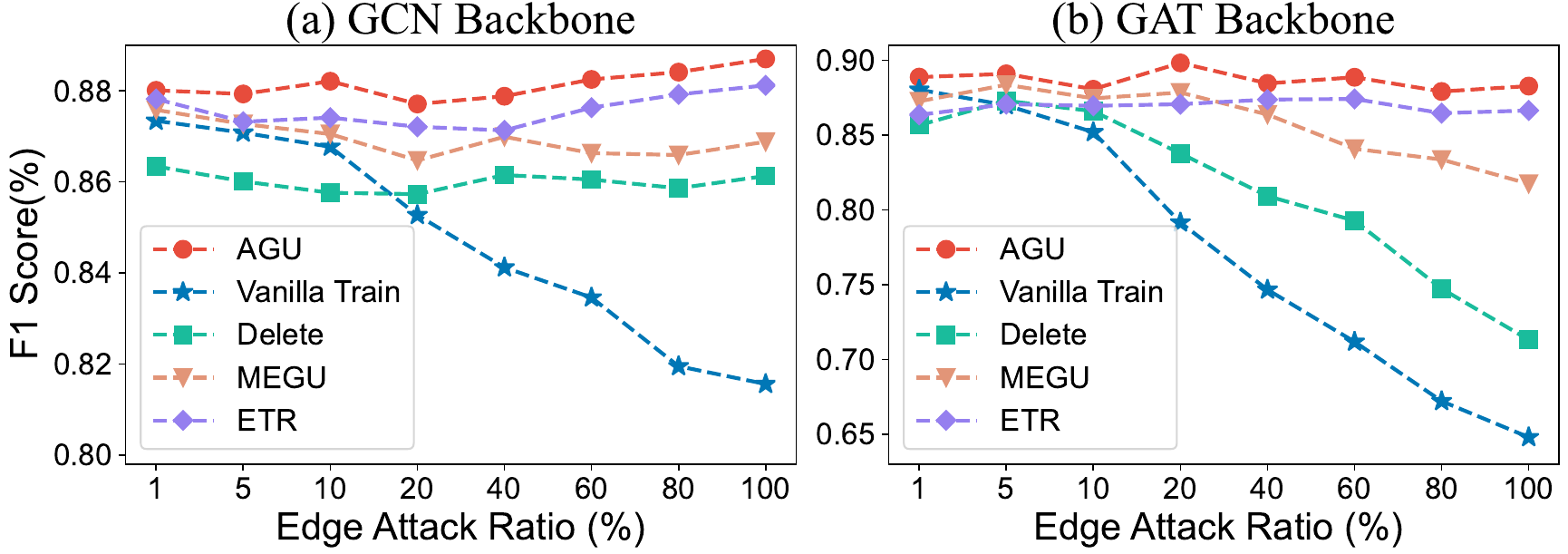}}
\caption{Edge attack performance on the Cora dataset.}
\label{atk_fig}
\end{figure}

\noindent\textbf{Efficiency.}
We report the average unlearning time in Table \ref{exp3}. The results show that AGU outperforms other training-based methods (Delete, MEGU, and Cognac), achieving a speedup from $1.36\times$ to $29\times$. Compared to training-free methods (GIF, IDEA, ETR), which directly adjust model parameters without training, AGU remains competitive, ranking second only to ETR in some cases. However, ETR requires additional time to extract subgraphs based on specific unlearning requests before initiating the unlearning process. Moreover, our experiments reveal that these training-free methods require careful parameter tuning to avoid model invalidation on the testing graph. In contrast, AGU ensures robust performance with minimal training epochs ($20$–$30$).

\noindent\textbf{Unlearning Capability.}
Fig. \ref{atk_fig} presents the node classification performance under different edge attack ratios. To provide a clear baseline for comparison, we introduce \textit{vanilla train}, which retrains a new GNN directly on the noisy graph. The results demonstrate that AGU consistently outperforms other baselines and remains stable across varying attack ratios, particularly when using GAT as the backbone GNN. Notably, unlike Delete, which adopts a random node-pair strategy for contrastive edge unlearning, AGU leverages homophily consistency between connected nodes to more effectively distinguish noisy edges from real ones, thereby mitigating the impact of edge attacks.

\subsection{Ablation Study (RQ2)}
We conduct ablation studies by creating five AGU variants to evaluate the contributions of its key components: (1) \textbf{w/o Homo:} This variant replaces the homophily-based node-pair strategy with a random node-pair strategy for contrastive edge unlearning, allowing us to assess the effectiveness of the proposed edge unlearning module. (2) \textbf{w/o EU} and \textbf{w/o FU:} Since AGU combines the edge unlearning (EU) and feature unlearning (FU) modules to address node unlearning tasks, these two variants independently evaluate the contributions of EU and FU to the overall unlearning performance. (3) \textbf{w/o MNF:} This variant removes the marginal neighbor filtering (MNF) module to evaluate its impact, particularly on degree-based GNNs (e.g., GCN). (4) \textbf{w/o ANS:} This variant excludes the affected neighbor selection (ANS) module to assess its influence on the unlearning process. The results in Table \ref{abl} show that AGU consistently outperforms all variants across all cases. This not only validates the importance of combining EU and FU for effective node unlearning, but also highlights the critical roles of MNF and ANS in identifying highly affected nodes for efficient and effective unlearning.

\begin{table}[t]
\setlength{\abovecaptionskip}{1pt}
\setlength{\belowcaptionskip}{0pt}
\centering
\caption{Node unlearning performance of AGU variants.}
\label{abl}
\resizebox{84mm}{21mm}{
\setlength{\tabcolsep}{2mm}{
\begin{tabular}{cc|ccccc}
\specialrule{0.05em}{1.5pt}{1.5pt}
Bone                  & Method  & Cora  & Citeseer & PubMed & Photo & Computer \\\specialrule{0.05em}{1.5pt}{1.5pt}
\multirow{6}{*}{GCN}  & AGU     & \textbf{85.94} & \textbf{76.22}    & \textbf{88.38}  & \textbf{92.61} & \textbf{86.34}    \\
& w/o Homo  & 85.52 & 74.28    & 87.74  & 92.38 & 85.61    \\
                      & w/o EU  & 85.07 & 73.81    & 87.52  & 92.06 & 85.48    \\
                      & w/o FU  & 84.73 & 74.89    & 86.88  & 91.94 & 85.37    \\
                      & w/o MNF & 85.41 & 75.34    & 87.69  & 92.13 & 85.62    \\
                      & w/o ANS & 85.29 & 74.26    & 87.44  & 91.82 & 85.75    \\\specialrule{0.05em}{1.5pt}{1.5pt}
\multirow{5}{*}{SAGE} & AGU     & \textbf{86.09} & \textbf{76.64}    & \textbf{90.34}  & \textbf{95.14} & \textbf{88.91}    \\
& w/o Homo  & 85.42 & 75.71    & 90.03  & 94.78 & 88.27    \\
                      & w/o EU  & 84.76 & 75.43    & 89.81  & 94.67 & 88.19    \\
                      & w/o FU  & 85.75 & 75.49    & 90.01  & 94.69 & 87.72    \\
                      & w/o ANS & 85.43 & 75.24    & 89.98  & 94.42 & 88.16   \\\specialrule{0.05em}{1.5pt}{1.5pt}
\end{tabular}}}
\end{table}

\subsection{Strategy Generalizability (RQ3)}
We evaluate the generalizability of our proposed neighbor selection strategies, MNF and ANS, by integrating them into existing GU methods: Delete, MEGU, and ETR. While MEGU and ETR have their own neighbor selection strategies, we replace them with MNF and ANS for comparison. In addition, we provide the “+AAN” variant, which selects accurate affected neighbors for degree-free GNNs (e.g., GAT). To reflect real-world scenarios with frequent but small-scale unlearning requests, we set the unlearn ratio to $1\%$. Our statistics show that in a 2-layer GNN, removing just $1\%$ of nodes affects over $70\%$ of the remaining nodes in the Photo and Computer datasets. We also conduct experiments on other datasets with various backbone GNNs and unlearn ratios (see Appendix). Our results demonstrate that MNF and ANS can be effectively integrated into existing GU frameworks, improving both effectiveness and efficiency across different datasets, backbone GNNs, and unlearn ratios. Specifically, the results in Table \ref{exp4} show that (1) For degree-based GNNs (e.g., GCN), applying MNF or ANS consistently reduces unlearning time while preserving prediction performance in most cases. Combining both strategies further improves prediction performance and reduces unlearning time by $11\%$ to $36\%$. (2) For degree-free GNNs (e.g., GAT), using ANS enhances prediction performance in most cases while achieving a comparable reduction in unlearning time ($25\%$ to $48\%$). 

\begin{table}[t]
\setlength{\abovecaptionskip}{1pt}
\setlength{\belowcaptionskip}{-2pt}
\centering
\caption{Performance improvement of baselines. Unlearn ratio=1\%.}
\label{exp4}
\resizebox{78mm}{38.7mm}{
\setlength{\tabcolsep}{3.3mm}{
\begin{tabular}{cc|cccc}
\specialrule{0.05em}{1pt}{1pt}
\multirow{2}{*}{Bone} & \multirow{2}{*}{Method} & \multicolumn{2}{c}{Photo}        & \multicolumn{2}{c}{Computer}     \\
                      &                         & F1               & Time          & F1               & Time          \\\specialrule{0.05em}{1pt}{1pt}
\multirow{12}{*}{GCN} & MEGU                    & 92.4±.3          & 0.34          & 85.8+.3          & 0.37          \\
                      & +MNF                    & 92.4±.7          & 0.29          & 85.1±.4          & 0.29          \\
                      & +ANS                    & 92.6±.6          & 0.30          & 85.6±.2          & 0.29          \\
                      & +Both                    & \textbf{92.9±.2} & \textbf{0.28} & \textbf{86.0±.3} & \textbf{0.28} \\ \cline{2-6} \rule{0pt}{9pt}
                      & Delete                  & 90.6±.3          & 1.85          & 83.9+.4          & 1.96          \\
                      & +MNF                    & 90.4±.7          & 1.81          & 84.3±.6          & 2.18          \\
                      & +ANS                    & 90.6±.5          & 1.79          & 84.9±.3          & 1.78          \\
                      & +Both                    & \textbf{90.8±.2} & \textbf{1.63} & \textbf{85.2±.3} & \textbf{1.62} \\ \cline{2-6} \rule{0pt}{9pt}
                      & ETR                     & 92.5±.3          & 0.07          & 85.8+.8          & 0.11          \\
                      & +MNF                    & 92.5±.8          & 0.07          & 85.2±.8          & 0.09          \\
                      & +ANS                    & 92.7±.4          & 0.07          & 85.8±.2          & 0.08          \\
                      & +Both                    & \textbf{92.8±.3} & \textbf{0.06} & \textbf{85.9±.3} & \textbf{0.07} \\\specialrule{0.05em}{1pt}{1pt}
\multirow{9}{*}{GAT}  & MEGU                    & 92.6±.4          & 0.48          & {86.0+.1} & 0.48          \\
                & +AAN                    & {92.7±.4} & {0.31} & \textbf{86.6±.5}          & {0.30} \\
                      & +ANS                    & \textbf{93.1±.3} & \textbf{0.27} & 86.4±.2          & \textbf{0.25} \\ \cline{2-6} \rule{0pt}{9pt}
                      & Delete                  & 90.9±.4          & 2.75          & 84.1+.3          & 2.74          \\
                & +AAN                    & {90.9±.3} & {2.24} & {84.2±.5} & {2.43} \\
                      & +ANS                    & \textbf{91.1±.4} & \textbf{1.98} & \textbf{84.7±.3} & \textbf{2.06} \\ \cline{2-6} \rule{0pt}{9pt}
                      & ETR                     & 92.2±.7          & 0.09          & {85.3+.3}          & 0.11          \\
                & +AAN                    & {92.2±.8} & {0.06} & {85.1±.8} & {0.08}\\
                      & +ANS                    & \textbf{92.7±.3} & \textbf{0.05} & \textbf{85.9±.5} & \textbf{0.07}\\\specialrule{0.05em}{1pt}{0pt}
\end{tabular}}}
\end{table}

\subsection{Parameter Study (RQ4)}  
\begin{figure}[t]
\setlength{\abovecaptionskip}{2pt}
\setlength{\belowcaptionskip}{-3pt}
\centering
\scalebox{0.292}{\includegraphics{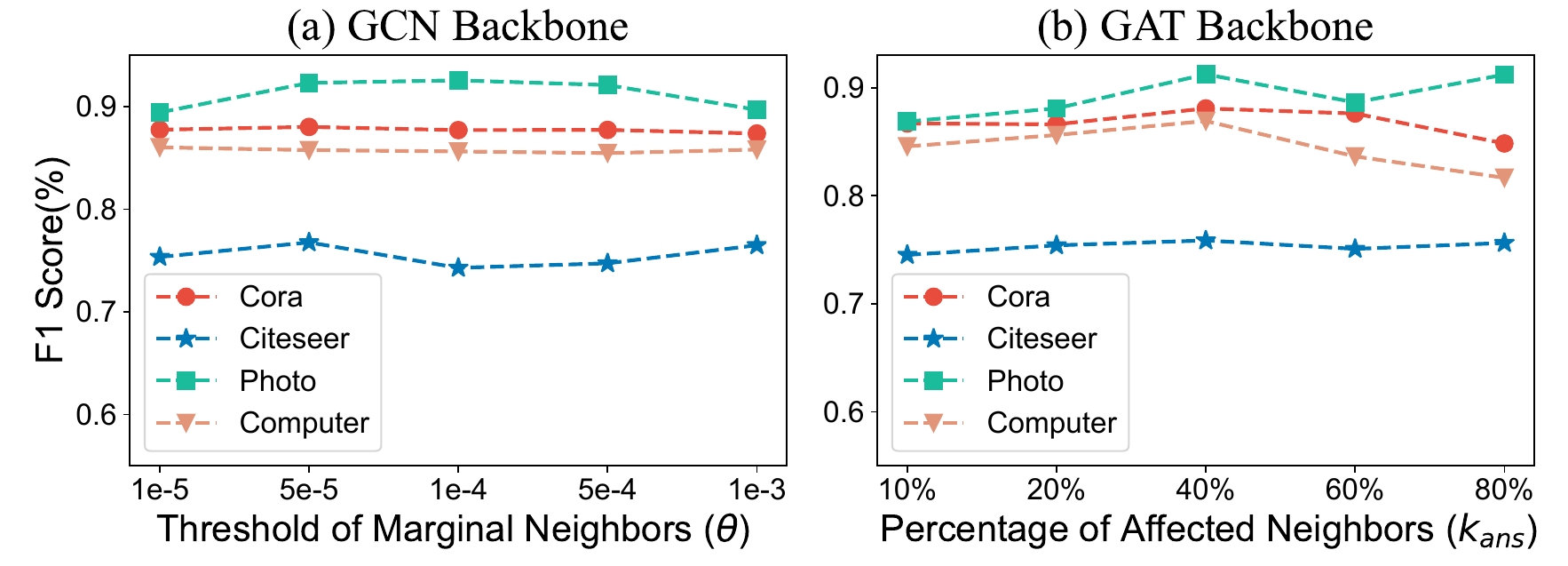}}
\caption{Node unlearning performance with varying parameters.}
\label{para}
\end{figure}

We investigate the sensitivity of two key parameters $\theta$ and $\emph{k}_\emph{ans}$ in AGU. The parameter $\theta$ controls the filtering of marginal neighbors ($\mathcal{N}_\emph{fmn}$) when using degree-based GNNs as backbones, while $\emph{k}_\emph{ans}$ determines the proportion of affected neighbors selected as highly affected neighbors ($\mathcal{N}_\emph{han}$). As shown in Fig. \ref{para}, increasing $\theta$ generally improves unlearning performance. However, when $\theta$ becomes too large, the number of filtered marginal neighbors approaches zero, leading to a performance degradation. Thus, the optimal range for $\theta$ is between $5e-5$ and $5e-4$. For $\emph{k}_\emph{ans}$, a value around $40\%$ can achieve a good trade-off between effectiveness and efficiency.

\section{Conclusion}
In this paper, we identify two significant limitations in existing GU studies: ineffective forgetting of unlearning elements and inaccurate identification of affected neighbors. These issues arise from the neglect of differences across various GU tasks and backbone GNNs. To address these issues, we propose AGU, an adaptive graph unlearning framework that unifies different GU tasks and develops general neighbor selection strategies adaptable to diverse backbone GNNs. Extensive experiments demonstrate the superior performance of AGU in terms of efficiency, effectiveness, and unlearning capability. In future work, we plan to extend our AGU framework to support various types of graph data, including heterogeneous graphs, knowledge graphs, and temporal graphs.

\section*{Acknowledgments}
This work is supported by the Australian Research Council Discovery Project DP230100676.

\bibliographystyle{named}
\bibliography{ijcai25}

\begin{thebibliography}{}

\bibitem[\protect\citeauthoryear{Chen \bgroup \em et al.\egroup }{2022}]{chen2022graph}
Min Chen, Zhikun Zhang, Tianhao Wang, Michael Backes, Mathias Humbert, and Yang Zhang.
\newblock Graph unlearning.
\newblock In {\em ACM CCS}, pages 499--513, 2022.

\bibitem[\protect\citeauthoryear{Chen \bgroup \em et al.\egroup }{2023}]{chen2023characterizing}
Zizhang Chen, Peizhao Li, Hongfu Liu, and Pengyu Hong.
\newblock Characterizing the influence of graph elements.
\newblock In {\em ICLR}, 2023.

\bibitem[\protect\citeauthoryear{Cheng \bgroup \em et al.\egroup }{2023}]{cheng2023gnndelete}
Jiali Cheng, George Dasoulas, Huan He, Chirag Agarwal, and Marinka Zitnik.
\newblock {GNND}elete: A general strategy for unlearning in graph neural networks.
\newblock {\em ICLR}, 2023.

\bibitem[\protect\citeauthoryear{Chien \bgroup \em et al.\egroup }{2023}]{chien2023efficient}
Eli Chien, Chao Pan, and Olgica Milenkovic.
\newblock Efficient model updates for approximate unlearning of graph-structured data.
\newblock In {\em ICLR}, 2023.

\bibitem[\protect\citeauthoryear{Cong and Mahdavi}{2022}]{cong2022grapheditor}
Weilin Cong and Mehrdad Mahdavi.
\newblock Grapheditor: An efficient graph representation learning and unlearning approach.
\newblock 2022.

\bibitem[\protect\citeauthoryear{Corso \bgroup \em et al.\egroup }{2024}]{corso2024graph}
Gabriele Corso, Hannes Stark, Stefanie Jegelka, Tommi Jaakkola, and Regina Barzilay.
\newblock Graph neural networks.
\newblock {\em Nature Reviews Methods Primers}, 4(1):17, 2024.

\bibitem[\protect\citeauthoryear{Ding \bgroup \em et al.\egroup }{2025}]{ding2025agu}
Pengfei Ding, Yan Wang, Guanfeng Liu, and Jiajie Zhu.
\newblock {AGU Appendix}.
\newblock \url{https://github.com/Aliezzz/AGU}, 2025.

\bibitem[\protect\citeauthoryear{Dong \bgroup \em et al.\egroup }{2024}]{dong2024idea}
Yushun Dong, Binchi Zhang, Zhenyu Lei, Na~Zou, and Jundong Li.
\newblock Idea: A flexible framework of certified unlearning for graph neural networks.
\newblock In {\em KDD}, pages 621--630, 2024.

\bibitem[\protect\citeauthoryear{Fan \bgroup \em et al.\egroup }{2025}]{fan2025opengu}
Bowen Fan, Yuming Ai, Xunkai Li, Zhilin Guo, Rong-Hua Li, and Guoren Wang.
\newblock {OpenGU}: A comprehensive benchmark for graph unlearning.
\newblock {\em arXiv preprint arXiv:2501.02728}, 2025.

\bibitem[\protect\citeauthoryear{Hamilton \bgroup \em et al.\egroup }{2017}]{hamilton2017inductive}
Will Hamilton, Zhitao Ying, and Jure Leskovec.
\newblock Inductive representation learning on large graphs.
\newblock {\em NeurIPS}, 30, 2017.

\bibitem[\protect\citeauthoryear{Khemani \bgroup \em et al.\egroup }{2024}]{khemani2024review}
Bharti Khemani, Shruti Patil, Ketan Kotecha, and Sudeep Tanwar.
\newblock A review of graph neural networks: concepts, architectures, techniques, challenges, datasets, applications, and future directions.
\newblock {\em Journal of Big Data}, 11(1):18, 2024.

\bibitem[\protect\citeauthoryear{Kipf and Welling}{2016}]{kipf2016semi}
Thomas~N Kipf and Max Welling.
\newblock Semi-supervised classification with graph convolutional networks.
\newblock {\em arXiv preprint arXiv:1609.02907}, 2016.

\bibitem[\protect\citeauthoryear{Kolipaka \bgroup \em et al.\egroup }{2024}]{kolipaka2024cognac}
Varshita Kolipaka, Akshit Sinha, Debangan Mishra, Sumit Kumar, Arvindh Arun, Shashwat Goel, and Ponnurangam Kumaraguru.
\newblock A cognac shot to forget bad memories: Corrective unlearning in gnns.
\newblock {\em arXiv preprint arXiv:2412.00789}, 2024.

\bibitem[\protect\citeauthoryear{Li \bgroup \em et al.\egroup }{2024a}]{li2024tcgu}
Fan Li, Xiaoyang Wang, Dawei Cheng, Wenjie Zhang, Ying Zhang, and Xuemin Lin.
\newblock {TCGU}: Data-centric graph unlearning based on transferable condensation.
\newblock {\em arXiv preprint arXiv:2410.06480}, 2024.

\bibitem[\protect\citeauthoryear{Li \bgroup \em et al.\egroup }{2024b}]{li2024towards}
Xunkai Li, Yulin Zhao, Zhengyu Wu, Wentao Zhang, Rong-Hua Li, and Guoren Wang.
\newblock Towards effective and general graph unlearning via mutual evolution.
\newblock In {\em AAAI}, pages 13682--13690, 2024.

\bibitem[\protect\citeauthoryear{Sekhari \bgroup \em et al.\egroup }{2021}]{sekhari2021remember}
Ayush Sekhari, Jayadev Acharya, Gautam Kamath, and Ananda~Theertha Suresh.
\newblock Remember what you want to forget: Algorithms for machine unlearning.
\newblock {\em NeurIPS}, 34:18075--18086, 2021.

\bibitem[\protect\citeauthoryear{Shchur \bgroup \em et al.\egroup }{2018}]{shchur2018pitfalls}
Oleksandr Shchur, Maximilian Mumme, Aleksandar Bojchevski, and Stephan G{\"u}nnemann.
\newblock Pitfalls of graph neural network evaluation.
\newblock {\em NeurIPS Workshop}, 2018.

\bibitem[\protect\citeauthoryear{Tan \bgroup \em et al.\egroup }{2024}]{tan2024unlink}
Jiajun Tan, Fei Sun, Ruichen Qiu, Du~Su, and Huawei Shen.
\newblock Unlink to unlearn: Simplifying edge unlearning in gnns.
\newblock In {\em TheWebConf}, pages 489--492, 2024.

\bibitem[\protect\citeauthoryear{Velickovic \bgroup \em et al.\egroup }{2017}]{velickovic2017graph}
Petar Velickovic, Guillem Cucurull, Arantxa Casanova, Adriana Romero, Pietro Lio, Yoshua Bengio, et~al.
\newblock Graph attention networks.
\newblock {\em stat}, 1050(20):10--48550, 2017.

\bibitem[\protect\citeauthoryear{Wang \bgroup \em et al.\egroup }{2023}]{wang2023inductive}
Cheng-Long Wang, Mengdi Huai, and Di~Wang.
\newblock Inductive graph unlearning.
\newblock In {\em USENIX Security}, pages 3205--3222, 2023.

\bibitem[\protect\citeauthoryear{Wu \bgroup \em et al.\egroup }{2019}]{wu2019simplifying}
Felix Wu, Amauri Souza, Tianyi Zhang, Christopher Fifty, Tao Yu, and Kilian Weinberger.
\newblock Simplifying graph convolutional networks.
\newblock In {\em ICML}, pages 6861--6871, 2019.

\bibitem[\protect\citeauthoryear{Wu \bgroup \em et al.\egroup }{2023a}]{wu2023gif}
Jiancan Wu, Yi~Yang, Yuchun Qian, Yongduo Sui, Xiang Wang, and Xiangnan He.
\newblock Gif: A general graph unlearning strategy via influence function.
\newblock In {\em TheWebConf}, pages 651--661, 2023.

\bibitem[\protect\citeauthoryear{Wu \bgroup \em et al.\egroup }{2023b}]{wu2023certified}
Kun Wu, Jie Shen, Yue Ning, Ting Wang, and Wendy~Hui Wang.
\newblock Certified edge unlearning for graph neural networks.
\newblock In {\em KDD}, pages 2606--2617, 2023.

\bibitem[\protect\citeauthoryear{Xu \bgroup \em et al.\egroup }{2018}]{xu2018powerful}
Keyulu Xu, Weihua Hu, Jure Leskovec, and Stefanie Jegelka.
\newblock How powerful are graph neural networks?
\newblock {\em arXiv preprint arXiv:1810.00826}, 2018.

\bibitem[\protect\citeauthoryear{Yang \bgroup \em et al.\egroup }{2016}]{yang2016revisiting}
Zhilin Yang, William Cohen, and Ruslan Salakhudinov.
\newblock Revisiting semi-supervised learning with graph embeddings.
\newblock In {\em ICLR}, pages 40--48, 2016.

\bibitem[\protect\citeauthoryear{Yang \bgroup \em et al.\egroup }{2025}]{yang2024erase}
Zhe-Rui Yang, Jindong Han, Chang-Dong Wang, and Hao Liu.
\newblock Erase then rectify: A training-free parameter editing approach for cost-effective graph unlearning.
\newblock In {\em AAAI}, pages 13044--13051, 2025.

\bibitem[\protect\citeauthoryear{Yi and Wei}{2025}]{yi2025scalable}
Lu~Yi and Zhewei Wei.
\newblock Scalable and certifiable graph unlearning: Overcoming the approximation error barrier.
\newblock In {\em ICLR}, 2025.

\bibitem[\protect\citeauthoryear{Zeng \bgroup \em et al.\egroup }{2019}]{zenggraphsaint}
Hanqing Zeng, Hongkuan Zhou, Ajitesh Srivastava, Rajgopal Kannan, and Viktor Prasanna.
\newblock Graphsaint: Graph sampling based inductive learning method.
\newblock In {\em ICLR}, 2019.

\bibitem[\protect\citeauthoryear{Zhu \bgroup \em et al.\egroup }{2020}]{zhu2020beyond}
Jiong Zhu, Yujun Yan, Lingxiao Zhao, Mark Heimann, Leman Akoglu, and Danai Koutra.
\newblock Beyond homophily in graph neural networks: Current limitations and effective designs.
\newblock {\em NeurIPS}, 33:7793--7804, 2020.

\end{thebibliography}

\end{document}